\documentclass{article}

\usepackage{microtype}
\usepackage{graphicx}
\usepackage{subfigure}
\usepackage{booktabs} 
\usepackage{amsmath,amssymb} 
\usepackage{makecell}
\usepackage{color}
\usepackage{enumitem}
\usepackage{multirow}
\usepackage{bbm}
\usepackage{arydshln}

\newcommand{\ie}{\textit{i}.\textit{e}., }
\newcommand{\eg}{\textit{e}.\textit{g}., }

\usepackage{hyperref}

\usepackage[accepted]{icml2020}

\icmltitlerunning{Operation-Aware Soft Channel Pruning using Differentiable Masks}

\begin{document}

\twocolumn[
\icmltitle{Operation-Aware Soft Channel Pruning using Differentiable Masks}



\icmlsetsymbol{equal}{*}

\begin{icmlauthorlist}
\icmlauthor{Minsoo Kang}{one}
\icmlauthor{Bohyung Han}{one}
\end{icmlauthorlist}

\icmlaffiliation{one}{Computer Vision Laboratory, Department of Electrical and Computer Engineering \& ASRI, Seoul National University, Korea}
\icmlcorrespondingauthor{Bohyung Han}{bhhan@snu.ac.kr}

\icmlkeywords{Machine Learning, ICML}

\vskip 0.3in
]

\printAffiliationsAndNotice{} 

\begin{abstract}
We propose a simple but effective data-driven channel pruning algorithm, which compresses deep neural networks in a differentiable way by exploiting the characteristics of operations.
The proposed approach makes a joint consideration of batch normalization (BN) and rectified linear unit (ReLU) for channel pruning; it estimates how likely the two successive operations deactivate each feature map and prunes the channels with high probabilities.
To this end, we learn differentiable masks for individual channels and make soft decisions throughout the optimization procedure, which facilitates to explore larger search space and train more stable networks.
The proposed framework enables us to identify compressed models via a joint learning of model parameters and channel pruning without an extra procedure of fine-tuning.
We perform extensive experiments and achieve outstanding performance in terms of the accuracy of output networks given the same amount of resources when compared with the state-of-the-art methods.
\end{abstract}

\section{Introduction}
\label{sec:introduction}
\begin{figure}[t!]
	\centering
	\includegraphics[width=0.96\linewidth]{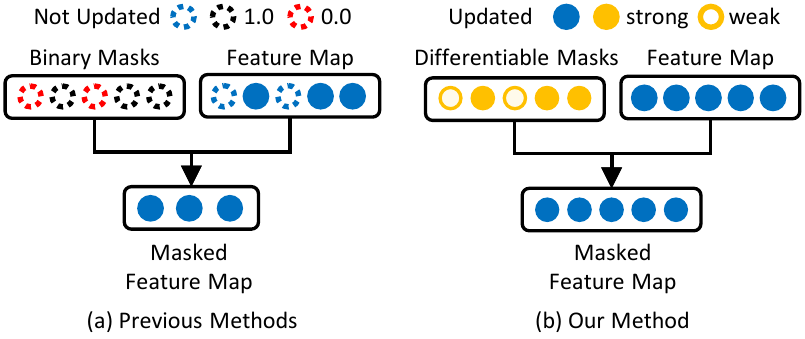}
	\vspace{-0.1cm}
	\caption{
		Illustration of the difference between the previous channel pruning methods and our approach.
		The previous methods often make hard decision for channel pruning and eliminate pruned channel permanently while our algorithm trains model parameters and masks jointly via soft channel pruning.
		Note that the pruned networks identified by the previous methods should be fine-tuned later but it may not be effective to achieve competitive accuracy because the pruned models are obtained from poor local optima due to greedy decisions during training procedure.  
		On the other hand, the proposed method achieves good performance without separate procedure of fine-tuning because we learn the model parameters and the pruning masks in a unified way based on a differentiable optimization framework.
		}
	\label{fig:intro}
\end{figure}

Deep neural networks have achieved state-of-the-art performance on various visual recognition tasks including image classification~\cite{he2016deep, huang2017densely, tan2019efficientnet}, image segmentation~\cite{noh2015learning, chen2018encoder}, object detection~\cite{he2017mask,li2019scale}, and visual tracking~\cite{nam2016learning}.
However, despite their outstanding accuracy, the applicability of the models based on deep neural networks to resource-hungry systems, \eg mobile or portable devices, is still limited  due to their computational or physical costs in terms of model sizes, FLOPs, and power consumption.
Consequently, model compression, a research problem to reduce the sizes of deep neural networks, has been investigated actively in the community. 

There is a large volume of research related to the compression of deep neural networks.
Early methods often rely on mathematical analysis of weight matrices such as matrix decomposition~\cite{denton2014exploiting, jaderberg2014speeding,tai2016convolutional}.
These approaches focus on reducing FLOPs of neural networks using the low-rank factorizations of pretrained weight matrices or tensors.
Another line of research in model compression is network quantization~\cite{courbariaux2015binaryconnect, han2016deep, rastegariECCV16, zhu2017trained, polino2018model}, which learns the low-bit representations of model parameters and is often employed for hardware integration of deep neural networks.
On the other hand, pruning approaches~\cite{han2015learning, molchanov2017variational, wen2016learning, li2017, luo2017thinet, liu2017learning, he2017channel, he2018soft, zhao2019variational, he2019filter} remove unnecessary weights or activations through heuristics, optimization techniques, or learning processes. 

We are interested in a structured pruning technique that removes channels from deep neural networks.
Most of the existing channel pruning methods rely on the two-step process for model compression---pruning followed by fine-tuning as illustrated in Figure~\ref{fig:intro}(a).
Since, at the time of fine-tuning, the networks are already pruned and have smaller capacity, the models may converge to poor local minima.
Moreover, the channel pruning decision of the previous methods often depends only on the absolute values of activations in the corresponding channels, but it fails to consider the natural consequences happening inside deep neural networks; negative activations will be discarded after passing rectified linear units (ReLUs) regardless of their magnitudes.

To deal with these issues, we incorporate training and pruning stages by introducing learnable mask parameters instead of identifying static masks.
To be specific, the model parameters and the pruning masks are learned jointly by a gradient-based optimization method, where the original capacity of the networks is preserved during training as illustrated in Figure~\ref{fig:intro}(b).
Furthermore, our framework considers the distribution of the activations in a channel together with the operations (batch normalization and ReLU) to be applied to the channel, and safely removes the channels that most of the values in the feature map are near zero or negative.
To this end, we propose a probabilistic formulation to identify such cases and estimate how likely individual channels are to be pruned by the criteria.

The main contributions of the proposed algorithm are summarized below:
\vspace{-0.2cm}
\begin{itemize}[label=$\bullet$]
	\item
	We propose a novel framework of channel pruning for deep neural network compression, which jointly optimizes model parameters and masks by employing a differentiable formulation.
	\item 
	We introduce a simple but effective channel pruning strategy, which estimates the importance of each channel probabilistically using the distribution of activations in accordance with the network operations.
	\item
	Experimental results show that the proposed approach outperforms the previous structured pruning methods.
	We also provide the empirical evidence that each component is helpful for improving accuracy.
\end{itemize}

The rest of the paper is organized as follows.
Section~\ref{sec:related} discusses the related work to model compression.
The details of our approach is described in Section~\ref{sec:framework}, and the experimental results are presented in Section~\ref{sec:experiments}.
Finally, we conclude this paper in Section~\ref{sec:conclusion}.

\section{Related Work}
\label{sec:related}

The proposed technique is one of the model compression algorithms, and aims to prune the channels that are unlikely to make a critical impact on the accuracy of the network.
This section first describes three types of model compression algorithms and then discusses several closely related works to our approach.

\subsection{Matrix Decomposition}
\label{sec:matrix decomposition}
The main goal of matrix decomposition methods is to reduce computational cost in a deep neural network by approximating weight matrices.
\cite{denton2014exploiting} approximates convolution filter weights to low-rank tensors by applying singular value decompositions.
\cite{jaderberg2014speeding} minimizes reconstruction error between the pretrained weights of original filters and a linear combination of basis filters while penalizing a nuclear norm for the approximated filters.
On the other hand, \cite{tai2016convolutional} reduces the reconstruction error of filters by minimizing the Frobenius norm of the difference between pretrained filters and approximated ones with a rank constraint, where they find a closed form solution.

\subsection{Network Quantization}
\label{sec:network quantization}
Network quantization techniques reduce the precision of weights to accelerate deep neural networks.   
\cite{courbariaux2015binaryconnect} learns a network with binary weights, where real-valued gradients are employed to update the binarized weights.
\cite{zhu2017trained} quantizes weights to \{$-W_l^n, 0, W_l^p$\} using learnable parameters, $W_l^n$ and $W_l^p$, which leads to a significantly smaller model with little accuracy drop.
In \cite{polino2018model}, the authors perform a model compression using quantization and knowledge distillation, where the quantization points are optimized through backpropagation.
Even though network quantization methods are effective to reduce computational cost conceptually, they require additional efforts in low-level processing to design practical systems. 

\subsection{Pruning}
\label{sec:pruning} 
Weight pruning methods eliminate unnecessary connections based on heuristics or optimization processes. 
Optimal Brain Damage~\cite{lecun1990optimal} removes weight parameters based on the Hessian matrix of objective function and fine-tunes the updated network.
\cite{han2015learning} prunes unimportant weights from a pretrained network based on the magnitude of the weights and retrain the model iteratively.
\cite{molchanov2017variational} proposes an extension of Variational Dropout~\cite{kingma2015variational}, which addresses the challenge in training with high dropout rates  and prunes the weights whose dropout rate is above a threshold.
Although weight pruning methods are successful in reducing a large number of connections, the resulting networks tend to have unstructured random connectivity, which leads to irregular memory access and little gain in actual inference speed without a proper hardware specialization as discussed in \cite{wen2016learning}.

Contrary to such pruning methods, structured pruning approaches~\cite{wen2016learning, li2017, luo2017thinet, liu2017learning, he2017channel, ye2018rethinking, he2018soft, huang2018data, zhao2019variational, he2019filter} aim to reduce redundant filters, channels or layers and improve the actual inference time without the need of special library or hardware support. 
\cite{li2017} removes unnecessary filters in a pretrained network based on their $\ell_1$ norms and fine-tunes the whole network after pruning.
Soft Filter Pruning (SFP)~\cite{he2018soft} resets less important filters at every epoch while updating all filters including the reset ones to identify to-be-pruned channels.
Sparse Structure Selection (SSS)~\cite{huang2018data} introduces scaling factors to prune specific structures such as neurons or residual blocks by adding sparsity regularizations on the structures.
Filter Pruning via Geometric Median (FPGM)~\cite{he2019filter} removes the filters minimizing the sum of distances to others instead of discarding the filters with negligible weights, as opposed to \cite{li2017,he2018soft}.
On the other hand, \cite{zhao2019variational} provides a Bayesian model compression technique, which approximates BN scaling parameters to a fully factorized normal distribution using stochastic variational inference~\cite{kingma2014auto} and then prunes the channels that have smaller mean and variance of the variational distribution.

\subsection{Channel Pruning without Extra Fine-Tuning}
The proposed algorithm is closely related to SFP~\cite{he2018soft}, SSS~\cite{huang2018data}, FPGM~\cite{he2019filter}, and Variational Pruning~\cite{zhao2019variational}  in the sense that they prune filters or channels without an extra fine-tuning stage.
Variational Pruning permanently removes some channels based on the predefined criteria at every epoch but such a greedy strategy may lead to a local optimum.
SFP and FPGM reset all the weights in unnecessary filters to zeros during training instead of completely eliminating the filters for their potential needs in the later stage of training.
However, the abrupt changes of the filter values make the training procedures unstable.
Both our algorithm and SSS introduce learnable masks and trains the masks and model parameters jointly.
The main difference between the two approaches is that ours estimates the masks based on BN and ReLU while SSS directly parameterizes and optimizes the masks by enforcing them to be zeros.

\section{Proposed Method}
\label{sec:framework}
This section describes our probabilistic framework of joint channel pruning and parameter optimization with differentiable deep neural network architectures in detail.  

\subsection{Preliminary}
\label{sec:preliminary}
Many recent deep neural networks~\cite{he2016deep, he2016identity, huang2017densely} adopt multiple identical building blocks that contain a batch normalization (BN) layer~\cite{ioffe2015batch} followed by a rectified linear unit (ReLU) layer.  
Originally, the BN layer has been designed to accelerate convergence and facilitate stable training of deep neural networks. 
The main idea of BN is to prevent the input feature distribution in each layer from fluctuating despite the inherent variations of data representations across mini-batches.
For the purpose, an output $x^{\text{out}}$ of a BN layer is calculated by normalizing an input $x^{\text{in}}$ and then performing an affine transformation as follows:
\begin{align} 
	z &=  \frac{x^{\text{in}} -\hat{\mu}}{\sqrt{\hat{\sigma}^2+\epsilon}}  \\
	x^{\text{out}} &= \gamma \cdot z + \beta
	\label{eq:bn}
\end{align}
where $\epsilon$ is a small positive value for numerical stability, and $\gamma$ and $\beta$ are learnable affine parameters in the BN layer.
Note that $\hat{\mu} \equiv \mathrm{E}[x^{\text{in}}]$ and $\hat{\sigma}^2 \equiv \mathrm{Var}[x^{\text{in}}]$ are given by respectively calculating the sample mean and variance sequentially across mini-batches in the training stage while they are fixed during testing. 
By employing the BN layer, we assume that $z$ follows a normal distribution, which makes $x^\text{out}$ normally distributed with mean $\beta$ and variance $\gamma^2$.

\begin{figure}[t!]
	\centering
	\includegraphics[width=0.96\linewidth]{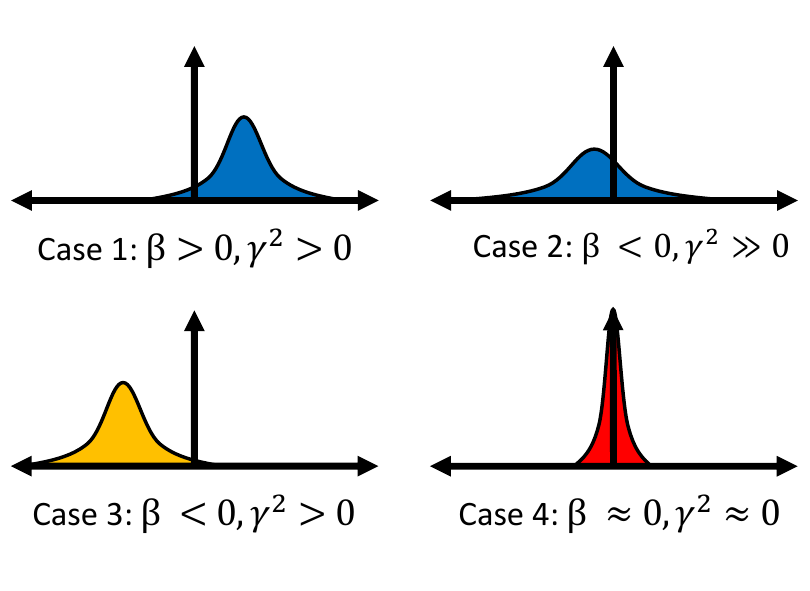}
	\vspace{-0.4cm}
	\caption{
		Illustration of the difference between the proposed method and the previous work~\cite{liu2017learning}.  
		In case 1 and 2, both approaches would not prune the channel due to their BN statistics.
		Our algorithm and \cite{liu2017learning} will make a different decision for case 3; the proposed method prunes the channel because most of the activations will be zeros after applying a ReLU function.
		The channel shown in case 4 will probably be pruned by both methods because the activations have low variance. 
		}
	\label{fig:gaussian}
	\vspace{0.5cm}
\end{figure}
%

\subsection{Mask for Channel Pruning}
\label{sub:mask}
\cite{liu2017learning} proposes a channel pruning method using BN layers, where they prune channels from a pretrained network if the learned scaling parameter $\gamma$ is less than a predefined threshold.  
This strategy is motivated by the fact that a channel such that the output $x^{\text{out}}$ is very close to a constant $\beta$ can be removed safely.
However, the performance gap between a pretrained network and the corresponding pruned network is often too high, which is partly because the method ignores the shifting parameters $\beta$ totally and relies on a heuristic hard channel pruning strategy as mentioned in \cite{zhao2019variational}. 

The proposed algorithm considers both the channel scaling and shifting parameters for channel pruning.
We claim that a feature map after a BN layer can be removed with low risk if most of its output activations are negative and then zeroed out by a subsequent ReLU layer eventually.
Figure~\ref{fig:gaussian} illustrates the difference between the prior work~\cite{liu2017learning} and the proposed approach; the channel pruning decision in \cite{liu2017learning} is based only on the $\gamma$ while our algorithm utilizes both the learnable parameters in a BN layer, $\beta$ and $\gamma$.
Note that the criteria in our approach take advantage of a subsequent ReLU layer and have low risk for pruning.

We define a mask variable $m(\delta;\beta, \gamma)$ given the predefined thresholds, $\delta$ and $c$, as
\begin{align} 
	m(\delta;\beta, \gamma ) &=
	\begin {cases}
	0 & \text{if~} \Phi(\delta ; \beta, \gamma) \geq c \\
	1 & \text{otherwise} 
	\end{cases},
	\label{eq:mask}
\end{align}
where $\Phi(\delta;\beta, \gamma)$ denotes the cumulative density function (CDF) of a Gaussian distribution $f(t ; \beta,\gamma)$ parametrized by $\beta$ and $\gamma$, which is given by
\begin{align}
\Phi(\delta ; \beta, \gamma) &= \int_{-\infty}^{\delta} f(t ; \beta,\gamma) \, dt \nonumber \\
&= \int_{-\infty}^{\delta}  \frac{1}{\sqrt{2\pi\gamma^2}} \exp \left(-\frac{(t-\beta)^2}{2\gamma^2} \right)  \, dt.
\end{align} 
The mask variable is not differentiable with respect to $\beta$ and $\gamma$, and the next subsection discusses how to make the function differentiable and how to optimize the masks and the model parameters jointly.

\subsection{Soft Channel Pruning}
\label{sub:soft}   
Our goal is to perform a joint optimization of model parameters and pruning masks by designing a differentiable deep neural network.
The model parameters can be updated by a gradient-based method trivially.
However, to learn the mask by the standard backpropagation, one can replace the mask function in \eqref{eq:mask}, which is based on an indicator function, by a logistic function, $q(\cdot)$, for its continuous relaxation as
\begin{align}
q(\delta ; \beta, \gamma) &= \frac{1}{1+\text{exp}(-k(\Phi(\delta ; \beta, \gamma) - c))}, 
\label{eq:mask_surrogate}
\end{align}
where $k$ is a constant.
Note that the logistic function becomes identical to the indicator function in \eqref{eq:mask} when $k$ approaches to infinity.
The partial derivative of the logistic function with respect to $\Phi(\delta ; \beta,\gamma)$ is given by
\begin{align}
 \frac{\partial q(\delta ; \beta,\gamma)}{\partial \Phi(\delta ; \beta,\gamma)} = k \cdot q(\delta ; \beta,\gamma) \cdot (1-q(\delta ; \beta,\gamma)),
 \label{eq:q}
\end{align}
which implies that, if $k$ is getting larger, the gradient vanishes over a wide range because $q(\delta ; \beta,\gamma)$ moves toward either 0 or 1 quickly.
Instead of setting $k$ to a large value, we consider a moderate value for $k$ to avoid the vanishing gradient problem.
Also, $q(\delta;\beta,\gamma)$ can be viewed as a probability mask to estimate how likely the corresponding channel is deactivated after going through a ReLU function.        
Note that $m(\cdot)$ is the \emph{discrete} mask function adopted for hard pruning at test time, which potentially has a large discrepancy from the probabilistic \emph{continuous} mask function $q(\cdot)$ employed during training time.

Directly sampling from a Bernoulli distribution is a reasonable option to tackle the issue, but the sampling procedure is not differentiable.
So, we employ the Gumbel-Softmax~\cite{jang2017categorical} trick, which performs a differentiable sampling to approximate to a categorical random variable.
Then, we define $n(\cdot)$ using the Gumbel-Softmax trick as
\begin{align}
	n&(\delta;\beta, \gamma) \\
	&= \frac{\text{exp}((\log \pi_1 + g_1) / \tau)}{\text{exp}((\log \pi_1 + g_1) / \tau) + \text{exp}((\log \pi_0 + g_0) / \tau)}, \nonumber
\end{align}
where $g_0$ and $g_1$ are samples drawn from $\text{Gumbel}(0,1)$ distribution, and $\pi_1$ and $\pi_0$ are given by $1-q(\delta;\beta,\gamma)$ and $q(\delta;\beta, \gamma)$, respectively.
Note that the output of $n(\cdot)$ becomes identical to a Bernoulli sample as $\tau$ approaches to 0.
To exploit the sample $n(\cdot)$ for training, the proposed algorithm revises a BN output $x^{\text{out}}$ to
\begin{align} 
	z &=  \frac{x^{\text{in}} -\hat{\mu}}{\sqrt{\hat{\sigma}^2+\epsilon}},  \\
	x^{\text{out}} &= (\gamma \cdot z + \beta) \cdot n(\delta ; \beta,\gamma).
	\label{eq:modified_bn}
\end{align}
In the revised BN layer,  the partial derivatives of $x^{\text{out}}$ with respect to $\beta$ and $\gamma$ are calculated as follows:
\begin{align}
 \frac{\partial x^{\text{out}}}{\partial \gamma} &= z \cdot n(\delta;\beta,\gamma) + (\gamma \cdot z + \beta ) \cdot \frac{\partial n(\delta;\beta,\gamma)}{\partial \gamma} \\
 \frac{\partial x^{\text{out}}}{\partial \beta} &=  n(\delta;\beta,\gamma) + (\gamma \cdot z + \beta ) \cdot \frac{\partial n(\delta;\beta,\gamma)}{\partial \beta} ,
\end{align}
where
\begin{align}
 \frac{\partial n(\delta;\beta,\gamma)}{\partial \gamma} &= \frac{\partial n(\delta;\beta,\gamma)}{\partial q(\delta;\beta,\gamma)} \cdot \frac{\partial q(\delta;\beta,\gamma)}{\partial \Phi(\delta;\beta, \gamma)} \cdot \frac{\partial \Phi(\delta;\beta, \gamma)}{\partial \gamma} \nonumber \\ 
 \frac{\partial n(\delta;\beta,\gamma)}{\partial \beta} &= \frac{\partial n(\delta;\beta,\gamma)}{\partial q(\delta;\beta,\gamma)} \cdot \frac{\partial q(\delta;\beta,\gamma)}{\partial \Phi(\delta;\beta, \gamma)} \cdot \frac{\partial \Phi(\delta;\beta, \gamma)}{\partial \beta} \nonumber
\end{align}
The derivatives of the above two terms also lead to the following equations:
\begin{align}
\frac{\partial n(\delta;\beta,\gamma)}{\partial q(\delta;\beta, \gamma)} &= -\frac{n(\delta;\beta, \gamma) (1-n(\delta;\beta, \gamma))}{\tau q(\delta;\beta, \gamma)(1-q(\delta;\beta, \gamma))} \\
\frac{\partial \Phi(\delta;\beta, \gamma)}{\partial \gamma} &= -f(\delta;\beta, \gamma) \cdot \frac{\delta-\beta}{|\gamma|} \cdot \frac{\partial |\gamma|}{\partial \gamma} ~\label{eq:cdf_derivative_gamma}\\ 
\frac{\partial \Phi(\delta;\beta, \gamma)}{\partial \beta} &=   -f(\delta;\beta, \gamma),
\end{align}
where $\frac{\partial q(\delta;\beta,\gamma)}{\partial \Phi(\delta;\beta,\gamma)} $ is given by \eqref{eq:q}.
Note that $\frac{\partial \Phi(\delta;\beta, \gamma)}{\partial \gamma}$ in \eqref{eq:cdf_derivative_gamma} is differentiable almost everywhere except for 0.
These equations illustrate that $\beta$ and $\gamma$ can be learned by the standard backpropagation. 

\subsection{Sparsity Loss with Confidence Interval}
\label{sub:sparsity}  
\begin{figure}[t!]
	\centering
	\includegraphics[width=1.00\linewidth]{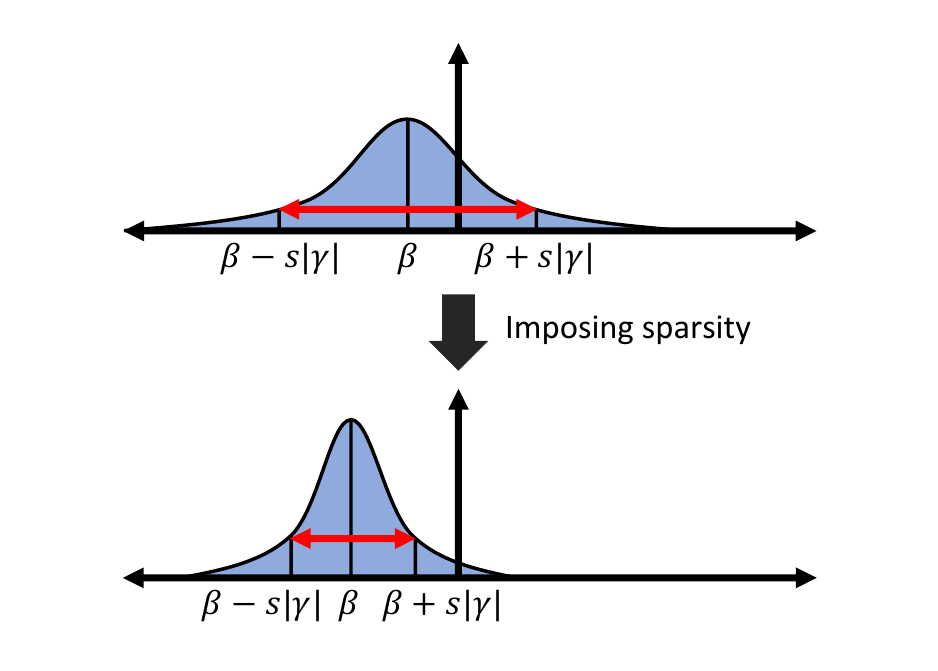}
	\vspace{-0.3cm}
	\caption{
		Illustration of confidence intervals in normal distributions, where $\beta$ and  
		$\gamma^2$ are mean and variance of the normal distribution, respectively, and $s$ is a constant. 
		The proposed sparsity regularization loss induces a larger value of CDF, $\Phi(\delta;\beta, \gamma)$ at the same threshold.
		}
	\label{fig:Sparsity_fig}
	\vspace{0.3cm}
\end{figure}
We have formulated the differentiable masks in Section~\ref{sub:mask} and \ref{sub:soft}, and the remaining question is how to define the loss function that makes networks compact by employing the differentiable masks.
We impose a sparsity regularization of channels using the mask function in \eqref{eq:mask}, which enforces $\Phi(\delta ; \beta, \gamma) \geq c$ for each channel.
The loss for the sparsity is given by
\begin{equation}
\mathcal{L}_\text{sparse}(\mathbb{B}, \mathbb{C}) = \sum_{\beta_{j} \in \mathbb{B}, \gamma_{j} \in \mathbb{C}}  \beta_j + s|\gamma_j|, 
\label{eq:sparsity_loss}
\end{equation}
where $\mathbb{B}$ and $\mathbb{C}$ are the sets of entire affine parameters in the BN layer in the original network and $s$ is a predefined constant. 
This loss is motivated by the confidence interval of the normal distribution, which is illustrated in Figure~\ref{fig:Sparsity_fig}.
Since $\delta$ is fixed, the sparsity loss maximizes the CDF, $\Phi(\delta ; \beta, \gamma)$.
Thus, the optimization based on this loss encourages the network to be more sparse.

As mentioned earlier, our algorithm performs a joint optimization of the model parameters and the pruning masks.
The full loss function $\mathcal{L}$ is expressed by a linear combination of the task-specific loss (classification loss in our paper) $\mathcal{L}_\text{cls}$ and the sparsity loss $\mathcal{L}_\text{sparse}$, which is given by
\begin{equation}
	\mathcal{L}(\mathbf{W}, \mathbb{B}, \mathbb{C}) = \mathcal{L}_\text{cls}(\mathbf{W}, \mathbb{B}, \mathbb{C}) + \lambda \, \mathcal{L}_{\text{sparse}}(\mathbb{B}, \mathbb{C}),
\end{equation}
where $\mathbf{W}$ is a set of all parameters in the network excluding the affine parameters in BN layers. 
The optimization based on the loss function is straightforward because the whole network is differentiable.
Note that we introduce no additional parameters to learn the masks.
Instead, the mask function in \eqref{eq:mask_surrogate} is determined by the parameters in BN layers, $\beta$ and $\gamma$.
Therefore, the gradient flow in the backward process for training is similar to the one in the ordinary deep neural networks with BN layers except that our approach performs the optimization of the model parameters and the masks jointly.

After training, we will obtain a deep neural network model for the target task (\ie classification) and a set of binary masks by thresholding CDF values based on $\mathbb{B}$ and $\mathbb{C}$.
The final network is given by applying the masks to the individual channels of the network.
Note that since the model parameters and pruning masks are learned at the same time, our algorithm does not require a separate fine-tuning stage followed by channel pruning.

%
\begin{table*}[!ht]
	\centering
	\caption{
		Performance comparison between the proposed algorithm, denoted by SCP, and other methods on CIFAR-10. 
	``FT" indicates whether the pruned network identified by each algorithm performs fine-tuning or not.
	``Baseline Acc." and ``Acc." mean the accuracies of the original and pruned networks, respectively.
	``Acc. Drop" represents the amount of accuracy degradation of the pruned networks with respect to the unpruned models.
	``Channels $\downarrow$", ``Param. $\downarrow$", and ``FLOPs $\downarrow$" denote the relative reductions in individual metrics compared to the unpruned networks.  
		A bold-faced number indicates the best performance in each category while ``--" denotes that the number is not available. 
	}
	\vspace{0.1in}
	\scalebox{0.75}{
		\begin{tabular}{c|c|c|c|c|c|c|c|c}		
			\toprule
			Method & Model & FT & Baseline Acc. (\%) & Acc. (\%) & Acc. Drop & Channels $\downarrow$ (\%) & Param. $\downarrow$ (\%) & FLOPs $\downarrow$ (\%) \\
			\hline \hline
			SFP~\cite{he2018soft} & ResNet-56   &X& 93.59 & 92.26 & 1.33 & 40  & --  & \textbf{52.60} \\
			FPGM~\cite{he2019filter} & ResNet-56 &X& 93.59 & 92.93 & 0.66 & 40  & --  & \textbf{52.60} \\
			SCP (Ours) & ResNet-56  & X & 93.69 & 93.23 & \textbf{0.46} & \textbf{45} & 48.47 &  51.50 \\
			\hline \hline
			Slimming~\cite{liu2017learning} & DenseNet-40&O& 94.39 & 92.59 & 1.80 & 80 & 73.53 &68.95 \\
			\hdashline
			Slimming~\cite{liu2017learning} & DenseNet-40&X& 94.39 & 12.78 & 81.61 & 80  & 73.53  & 68.95 \\
			Variational Pruning~\cite{zhao2019variational} &DenseNet-40 & X & 94.11 & 93.16 & 0.95 & 60 & 59.76 & 44.78 \\
			SCP (Ours) & DenseNet-40  & X & 94.39 & 93.77 & \textbf{0.62} & \textbf{81} & \textbf{75.41} & \textbf{70.77} \\ 
			\hline \hline
			Slimming~\cite{liu2017learning} & VGGNet-19&O& 93.84 & 93.21 & 0.63 & 80 & 93.10 & 63.39 \\
			\hdashline
			Slimming~\cite{liu2017learning} & VGGNet-19&X& 93.84 & 10.00 & 83.84 & 80 & 93.10 & 63.39 \\
			SCP (Ours) & VGGNet-19  & X & 93.84 & 93.82 & \textbf{0.02} & \textbf{81} & \textbf{95.21} & \textbf{74.06} \\
			\hline \hline
			Slimming ~\cite{liu2017learning}& VGGNet-16&O& 93.85 & 92.91 & 0.94 & 70 & 87.97 & 48.12 \\
			\hdashline
			Slimming~\cite{liu2017learning}& VGGNet-16&X& 93.85 & 10.00 & 83.85 & 70 & 87.97 & 48.12 \\		
			Variational Pruning~\cite{zhao2019variational}&VGGNet-16 & X & 93.25 & 93.18 & 0.07 & 62 &73.34 & 39.10 \\
			SCP (Ours) & VGGNet-16  & X & 93.85 & 93.79 & \textbf{0.06} & \textbf{75} & \textbf{93.05} & \textbf{66.23} \\
			\bottomrule
		\end{tabular}
	}
	\label{table:cifar10_table}
	\vspace{0.1cm}
\end{table*}
%

\section{Experiments}
\label{sec:experiments}
This section first presents details of datasets and our implementation, and then discusses the performance of the proposed algorithm in comparison to existing methods.
We also analyze the characteristics of our approach via various ablative experiments.

\subsection{Dataset}
\label{sec:dataset}
We employ CIFAR-10/100 and ILSVRC-12 in our experiment, which are the datasets widely accepted for evaluation of model compression techniques. 
The CIFAR-10/100 datasets consist of 50K and 10K color image splits for training and testing, where the size of each image is $32 \times 32$.
The only difference between the two editions is the number of classes; CIFAR-10 and CIFAR-100 have 10 and 100 classes, respectively.
On the other hand, the ILSVRC-12 dataset~\cite{ILSVRC15} contains 1,281,167 training images in color and 50,000 color images for validation in 1,000 classes.
For preprocessing images, we follow the techniques in \cite{liu2017learning} and \cite{he2019filter} for CIFAR-10/100 and ILSVRC-12, respectively, to make fair comparisons.

\subsection{Implementation Details}
\label{sec:imple_details}
The proposed method is implemented using TensorFlow library~\cite{tensorflow2015-whitepaper}.
We train the network using SGD with Nesterov momentum~\cite{sutskever2013importance} 0.9, weight decay parameter 0.0001, and initial learning rate 0.1.
The setting for the experiment on the CIFAR datasets follows the one used in \cite{liu2017learning}, where the batch size is set to 64, and the learning rate is reduced by the factor of 10 after the 80$^\text{th}$ and 120$^\text{th}$ epochs.
For the ILSVRC-12 dataset, the network is learned for 100 epochs with 4 GPUs, where the total batch size is 256 and the  learning rate is dropped by a factor of 10 at the 30$^\text{th}$, 60$^\text{th}$, and 90$^\text{th}$ epochs.
Fine-tuning the pruned network is based on the same setting except for the initial learning rate of 0.01.
We set the temperature parameter $\tau$ for Gumbel-Softmax~\cite{jang2017categorical} to 0.5 and the threshold $\delta$ for CDF to 0.05 for all pruned models. 
All networks are trained from scratch.

\subsection{Results on CIFAR-10/100 Datasets}
\label{sec:results_cifar}
We evaluate the performance of the proposed algorithm, denoted by SCP (Soft Channel Pruning), on the CIFAR-10/100 datasets using ResNet~\cite{he2016deep}, DenseNet~\cite{huang2017densely}, and VGGNet~\cite{simonyan2015very} since they are widely used for image recognition tasks.
We compare our framework with Slimming~\cite{liu2017learning} and Variational Pruning~\cite{zhao2019variational}, which also take advantage of BN layers to prune redundant channels. For CIFAR-10,  we also compare the proposed method with SFP~\cite{he2018soft} and FPGM~\cite{he2019filter}, which do not require fine-tuning like our algorithm.  
The results of Slimming are given by our reproduction from TensorFlow implementation.

\begin{table*}[!t]
	\centering
	\caption{
		Performance of our algorithm, SCP, with respect to Slimming and Variational Pruning on CIFAR-100. 
	}
	\vspace{0.1in}
	\scalebox{0.75}{
		\begin{tabular}{c|c|c|c|c|c|c|c|c}		
			\toprule
			Method & Model & FT & Baseline Acc. (\%) & Acc. (\%) & Acc. Drop & Channels $\downarrow$ (\%) & Param. $\downarrow$ (\%) & FLOPs $\downarrow$ (\%) \\
			\hline \hline
			Slimming~\cite{liu2017learning}& ResNet-164&O& 77.24 & 74.52 & 2.72 &\textbf{60} & \textbf{29.26} & \textbf{47.92} \\
			\hdashline
			SCP (Ours) & ResNet-164  & X & 77.24 & 76.62 & \textbf{0.62} & 57 & 28.89 & 45.36 \\
			\hline \hline
			Slimming~\cite{liu2017learning}& DenseNet-40&O& 74.24 & 73.53 & 0.71 & \textbf{60} & 54.99 & \textbf{50.32} \\
			\hdashline
			Variational Pruning~\cite{zhao2019variational} &DenseNet-40 & X & 74.64 & 72.19 & 2.45 & 37 & 37.73 & 22.67 \\
			SCP (Ours)  & DenseNet-40  & X & 74.24 & 73.84 & \textbf{0.40} & \textbf{60} & \textbf{55.22} & 46.25 \\ 
			\hline \hline
			Slimming ~\cite{liu2017learning}& VGGNet-19&O& 72.56 & 73.01 & \textbf{-0.45} &  50 & 76.47 & 38.23 \\
			\hdashline
			SCP (Ours) & VGGNet-19  & X & 72.56 & 72.99 & -0.43 & \textbf{51} & \textbf{77.52} & \textbf{40.92} \\
			\hline \hline
			Slimming ~\cite{liu2017learning} & VGGNet-16&O& 73.51 & 73.45 & 0.06 & 40 & 66.30 & 27.86 \\
			\hdashline
			Variational Pruning~\cite{zhao2019variational} &VGGNet-16 & X & 73.26 & 73.33 & -0.07 & 32 &37.87 & 18.05 \\
			SCP (Ours) & VGGNet-16  & X & 73.51 & 73.86 & \textbf{-0.35} & \textbf{52} & \textbf{80.14} & \textbf{51.45} \\
			\bottomrule
		\end{tabular}
	}
	\label{table:cifar100_table_1}
\end{table*}
\begin{table*}[!t]
	\centering
	\caption{
		Performance of our algorithm, SCP, with respect to Slimming on CIFAR-100 with high pruning ratios. 
	}
	\vspace{0.1in}
	\scalebox{0.75}{
		\begin{tabular}{c|c|c|c|c|c|c|c|c}		
			\toprule
			Method & Model & FT & Baseline Acc. (\%) & Acc. (\%) & Acc. Drop & Channels $\downarrow$ (\%) & Param. $\downarrow$ (\%) & FLOPs $\downarrow$ (\%) \\
			\hline \hline
			Slimming ~\cite{liu2017learning} & ResNet-164&O& 77.24 & 71.54 & 5.70 & 70& 40.72 & 62.29 \\
			\hdashline
			SCP (Ours) & ResNet-164  & X & 77.24 & 75.05 & \textbf{2.19} & \textbf{71} & \textbf{53.30} & \textbf{64.93} \\
			\hline \hline
			Slimming~\cite{liu2017learning}& DenseNet-40&O& 74.24 & 70.97 & 3.27 & \textbf{80} & 73.62 & \textbf{68.20} \\
			\hdashline
			SCP (Ours)  & DenseNet-40  & X & 74.24 & 73.17 & \textbf{1.07} & \textbf{80} & \textbf{74.86} & 67.82 \\ 
			\hline \hline
			Slimming ~\cite{liu2017learning} & VGGNet-19&O& 72.56 & 67.81 & 4.75 & \textbf{70} & 89.10 & 59.67 \\
			\hdashline
			SCP (Ours) & VGGNet-19  & X & 72.56 & 72.15 & \textbf{0.41} & 67 & \textbf{89.37} & \textbf{61.94} \\
			\hline \hline
			Slimming~\cite{liu2017learning} & VGGNet-16&O& 73.51 & 65.22 & 8.29 & 70 & 92.46 & \textbf{80.49} \\
			\hdashline
			SCP (Ours) & VGGNet-16  & X & 73.51 & 69.96 & \textbf{3.55} & \textbf{72} & \textbf{94.01} & 79.24 \\
			\bottomrule
		\end{tabular}
	}
	\label{table:cifar100_table_2}
\end{table*}
%
\begin{table*}[!t]
	\centering
	\caption{
		Performance of our algorithm, SCP, with respect to SSS and FPGM on ILSVRC-12 using ResNet-50. 
	}
	\vspace{0.1in}
	\scalebox{0.75}{
		\begin{tabular}{c|c|c|c|c|c|c}
			\toprule
			Method & FT & Baseline Top-1 Acc. (\%) & Baseline Top-5 Acc. (\%) & Top-1 Acc. Drop & Top-5 Acc. Drop & FLOPs $\downarrow$ (\%) \\
			\hline\hline	
			FPGM~\cite{he2019filter} & O & 76.15 & 92.87 & 1.32 & \textbf{0.55} & 53.5 \\
			SCP (Ours) & O & 75.89 & 92.98 & \textbf{0.62} & 0.68 & \textbf{54.3}\\
			\hdashline
			SSS~\cite{huang2018data} & X & 76.12 & 92.86 & 4.30 & 2.07 & 43.0 \\ 
			FPGM~\cite{he2019filter}  & X &76.15 & 92.87  & 2.02 & \textbf{0.93} & 53.5 \\
			SCP (Ours) & X & 75.89 & 92.98 & \textbf{1.69} & 0.98 & \textbf{54.3} \\
			\bottomrule
			
		\end{tabular}
	}
	\label{table:ImageNet_table}
\end{table*}
%

\paragraph{CIFAR-10}
Table~\ref{table:cifar10_table} presents that the proposed approach, SCP, achieves the lowest accuracy drops compared to SFP, FPGM, 
Slimming, and Variational Pruning in all tested scenarios regardless of backbone networks.
Also, the pruned models given by SCP based on DenseNet-40, VGGNet-19, and VGGNet-16 outperform the ones identified by Slimming with fine-tuning by 1.18\%, 0.61\%, and 0.88\% points, respectively, although our models do not go through separate fine-tuning procedures.
As mentioned in Section~\ref{sub:mask}, Slimming suffers from significant accuracy drops without fine-tuning because of the ignorance of the shift parameter $\beta$ in BN layers and the heuristic channel removal incurring error propagation over the network.
In the case of VGGNet-19 and VGGNet-16, the pruned models identified by our algorithm have the almost same accuracy with the baselines while achieving substantial speed-ups. 
In addition, our method presents comparable or superior accuracy compared to the method based on Variational Pruning even with significantly smaller network sizes when DensNet-40 and VGGNet-16 are used as backbone networks.

\paragraph{CIFAR-100}
Table~\ref{table:cifar100_table_1} shows that our models given by SCP achieve less accuracy drop compared to the pruned networks identified by Slimming and Variational Pruning in most cases.
Specifically, in the case of DenseNet-40 and VGGNet-16, our algorithm outperforms Variational Pruning by 1.95\% and 0.28\% points, respectively, even though our models are more compact.
In addition, SCP achieves outstanding performance compared to Slimming on ResNet-164, DenseNet-40, and VGGNet-16 in terms of accuracy drop while the proposed method is comparable to Slimming on VGGNet-19.
To demonstrate the effectiveness of our algorithm further, we also present the results with high pruning ratios in Table~\ref{table:cifar100_table_2}. 

\subsection{Results on ILSVRC-12}
\label{sec:experimental on imagenet}
We compare the proposed method with SSS~\cite{huang2018data} and FPGM~\cite{he2019filter} on a large-scale dataset, ILSVRC-12, and present the results in Table~\ref{table:ImageNet_table}.
SCP accomplishes outstanding results compared to SSS for all measures and comparable performance to FPGM. 
Especially, the pruned model given by our approach without fine-tuning outperforms the one identified by SSS without fine-tuning by 2.61\% and 1.09\% points in top-1 and top-5 accuracy drop, respectively, even though our model is smaller. 
Also, SCP exceeds FPGM by 0.70\% and 0.33\% points in the top-1 accuracy drop with and without fine-tuning, respectively.
Although our approach is marginally worse than FPGM in terms of top-5 accuracy drop, it has higher compression rates.

To observe the realistic and practical speed-up of the proposed method, we measure the wall clock inference time for the unpruned and pruned models on the NVIDIA TITAN Xp with a batch size of 64. 
SCP achieves about $24\%$ reduction of inference time, from 104 ms without pruning to 79 ms with pruning.

\subsection{Analysis}
\label{analysis}

\paragraph{Consideration of ReLU for channel pruning}
One of the main ideas in this paper is to take advantage of BN and ReLU together for channel pruning, which is clearly differentiated from Slimming~\cite{liu2017learning}, which is based only on BN.
So, we first analyze how critical ReLU in our framework by testing performance on CIFAR-100 with ResNet-164, DenseNet-40, VGGNet-19, and VGGNet-16.
For the purpose, we evaluate a modified version of our algorithm with differential masks under consideration of BN operation only, which makes our pruning criteria similar to Slimming~\cite{liu2017learning}. 
In other words, the CDF $\Phi(\delta;\beta, \gamma)$ is replaced by $P(-\delta_{\text{new}} \leq x^{\text{out}} \leq \delta_{\text{new}})$, where $\delta_{\text{new}}$ is a small positive number; if the most activations in a channels are near-zeros, the channel is to be pruned in the modified formulation.
In addition, we makes the objective function more appropriate for the new formulation by revising the sparsity loss in \eqref{eq:sparsity_loss} to the following one,
\begin{equation}
\mathcal{L}_\text{sparse}(\mathbb{B}, \mathbb{C}) = \hspace{-0.3cm} \sum_{\beta_{j} \in \mathbb{B}, \gamma_{j} \in \mathbb{C}} |\beta_j + s|\gamma_j|| + |\beta_j - s|\gamma_j||.
\label{eq:sparsity_loss_revised}
\end{equation}

Table~\ref{table:ReLU_ablation} illustrates that our full model outperforms the modified version, denoted by ``SCP without ReLU".
Note that SCP without ReLU is still better than Slimming in terms of accuracy drop even though their criteria for channel pruning are similar to each other.
The results highlight two advantages of our algorithm; 1) the consideration of BN and ReLU together for channel pruning is effective, and 2) our joint optimization framework with differentiable masks drives the pruned network to converge to a better model compared to Slimming, which determines masks by a heuristic.

\begin{table}[!t]
	\centering
	\caption{
	 Results with and without consideration of ReLU operations on CIFAR-100.	
	 ``SCP without ReLU" indicates that only BN is employed for channel pruning decisions and we only prune the channels when the absolute value of the channels are below a threshold (\eg case 4 in Figure~\ref{fig:gaussian}). 
	}
	\vspace{0.1in}
	\scalebox{0.75}{
		\begin{tabular}{c|c|c|c}
			\toprule
			 Method & Network &Acc. Drop & FLOPs $\downarrow$ (\%) \\
			\hline\hline	
			SCP (Ours) & ResNet-164 & \textbf{0.62} & 45.36 \\
			SCP without ReLU & ResNet-164 & 1.33 & \textbf{48.89} \\
			Slimming~\cite{liu2017learning}& ResNet-164& 2.72 & 47.92 \\
			\hline \hline
			SCP (Ours)  & DenseNet-40 & \textbf{0.77} & 65.49 \\
			SCP without ReLU& DenseNet-40 & 1.79 & 62.31 \\
			Slimming~\cite{liu2017learning}& DenseNet-40& 3.27 & \textbf{68.20} \\
			\hline \hline
			SCP (Ours) & VGGNet-19 & \textbf{0.41} & \textbf{61.94} \\
			SCP without ReLU & VGGNet-19 & 1.33 & 54.96 \\
			Slimming ~\cite{liu2017learning} & VGGNet-19& 4.75 & 59.67 \\
			\hline \hline
			SCP (Ours) & VGGNet-16 & \textbf{3.55} & 79.24 \\
			SCP without ReLU & VGGNet-16 & 5.21 & 78.37 \\
			Slimming~\cite{liu2017learning} & VGGNet-16 & 8.29 & \textbf{80.49} \\
			\bottomrule			
		\end{tabular}
	}
	\label{table:ReLU_ablation}
\end{table}
\begin{figure*}[t]
\centering
    \subfigure[ResNet-164]{\label{fig:4b}\includegraphics[width=42mm, trim=0 200 0 200, clip] {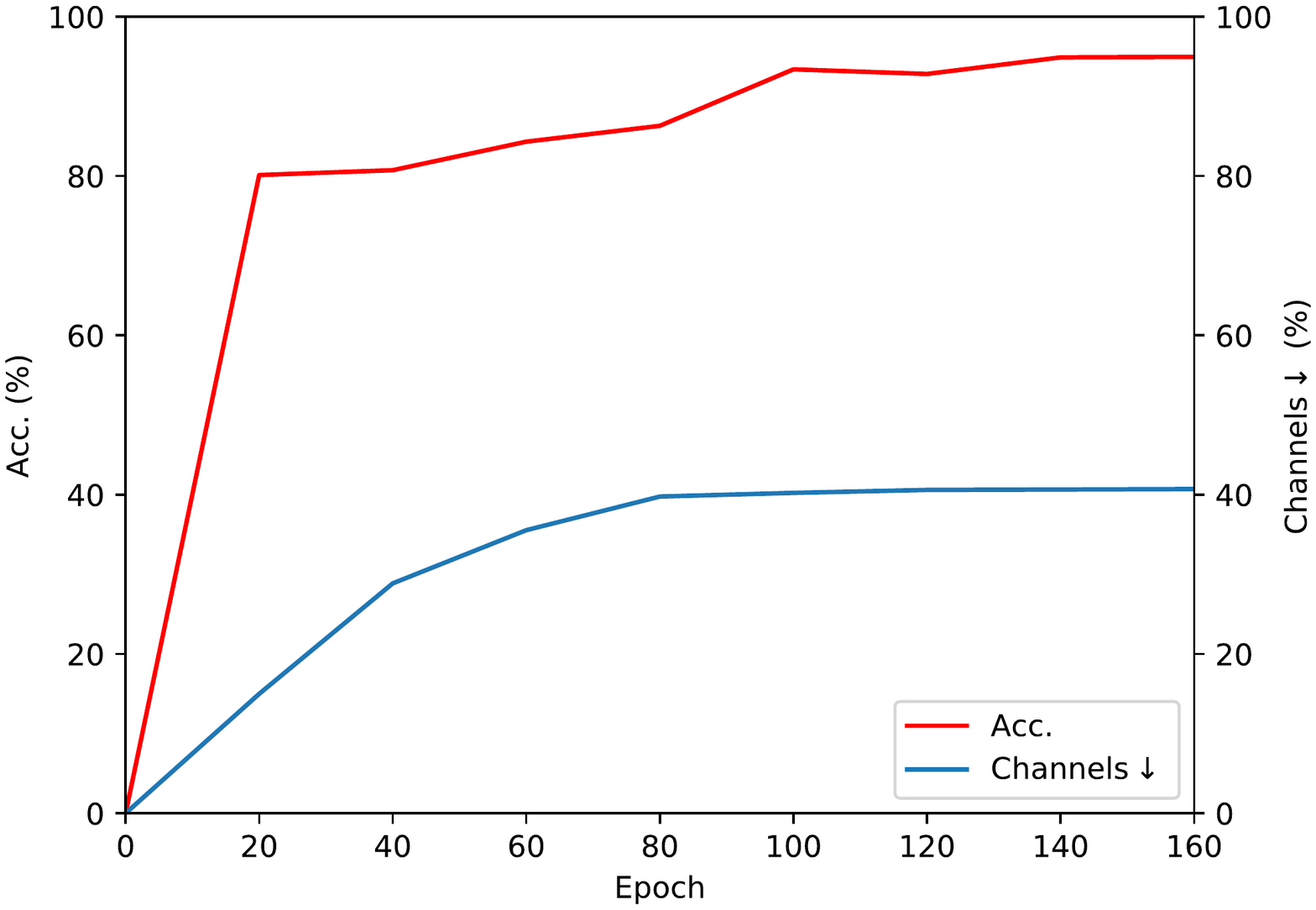}}
    \subfigure[DenseNet-40]{\label{fig:4a}\includegraphics[width=42mm, trim=0 200 0 200, clip] {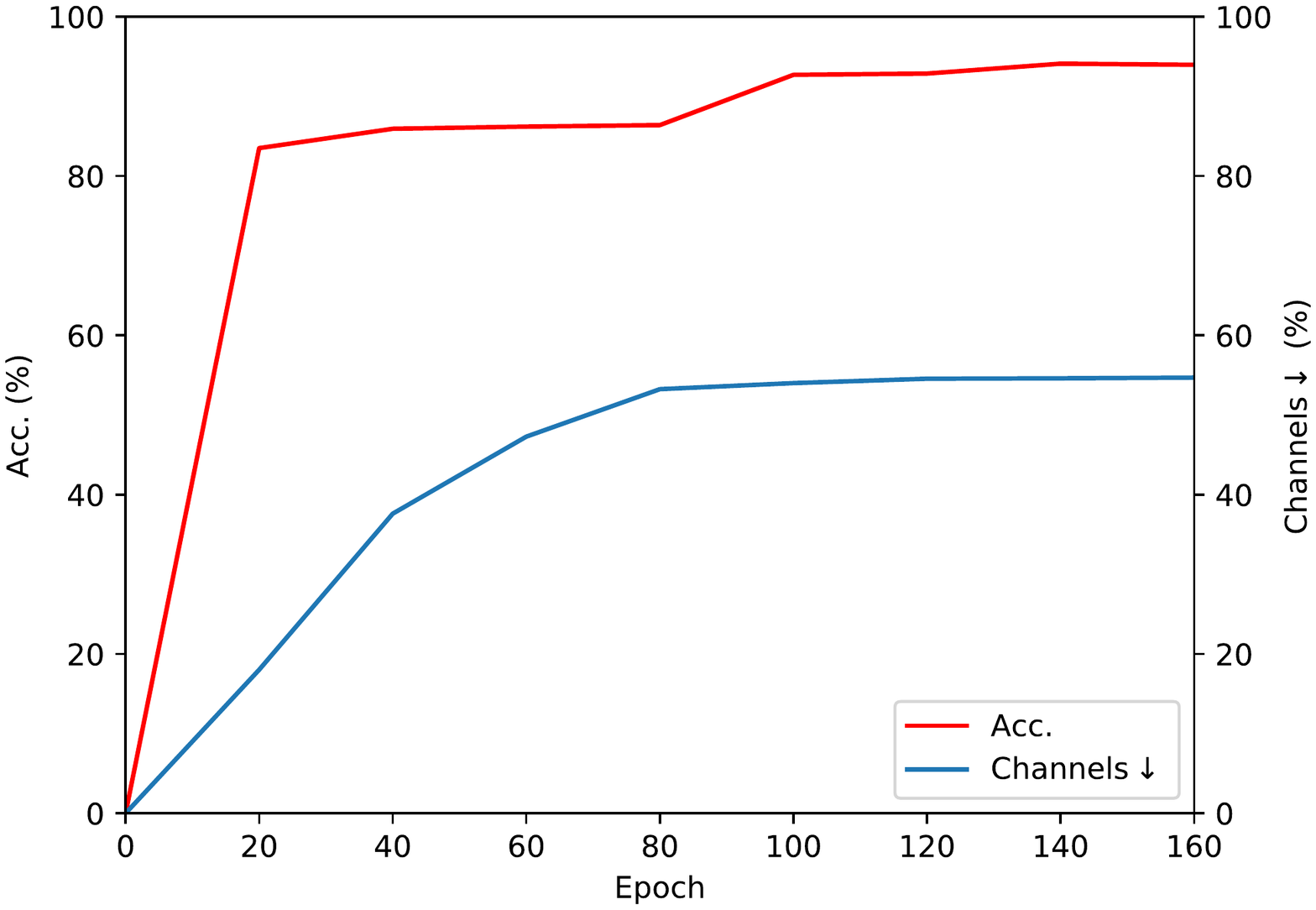}}
    \subfigure[VGGNet-19]{\label{fig:4c}\includegraphics[width=42mm, trim=0 200 0 200, clip] {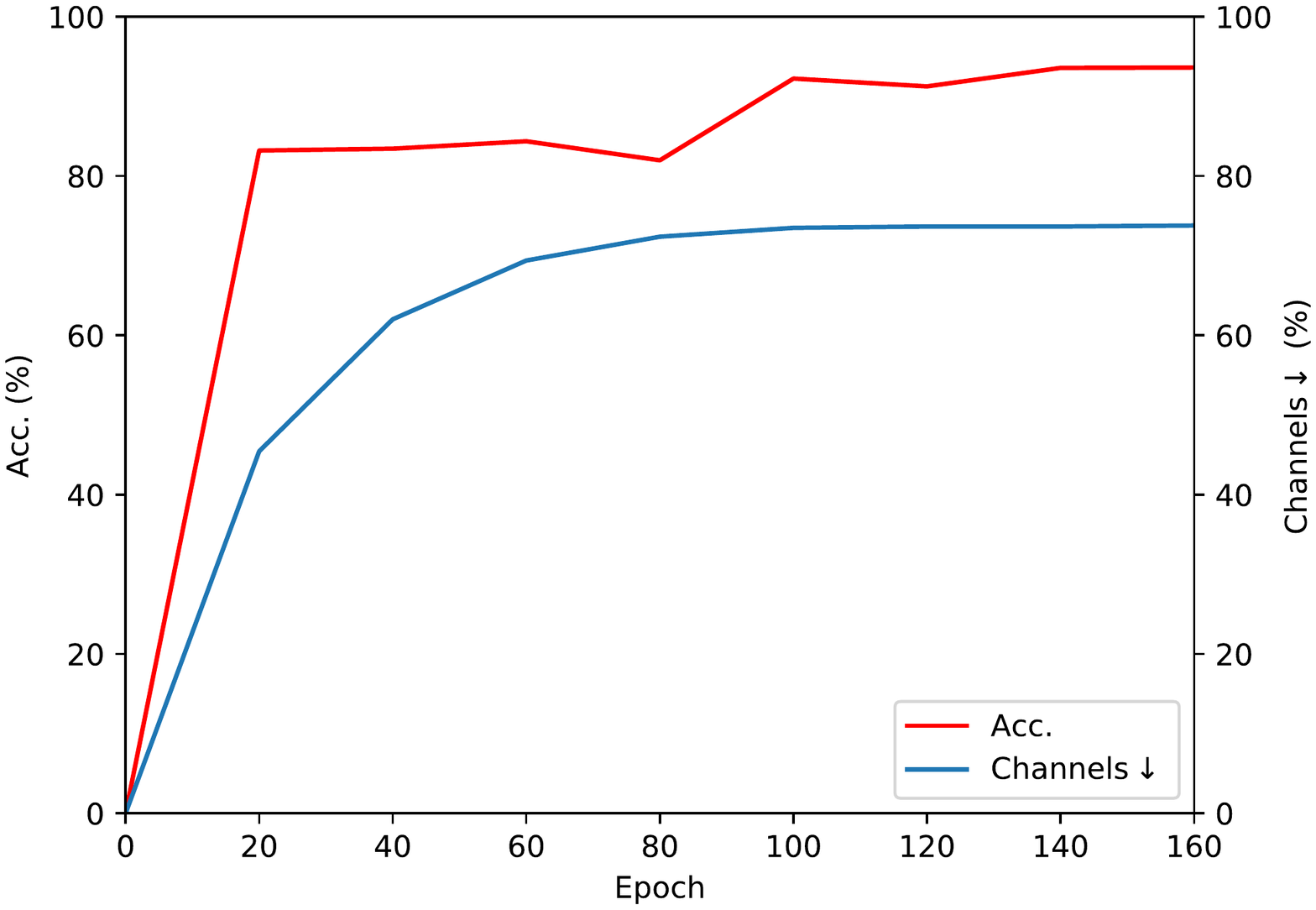}}
    \subfigure[VGGNet-16]{\label{fig:4d}\includegraphics[width=42mm, trim=0 200 0 200, clip] {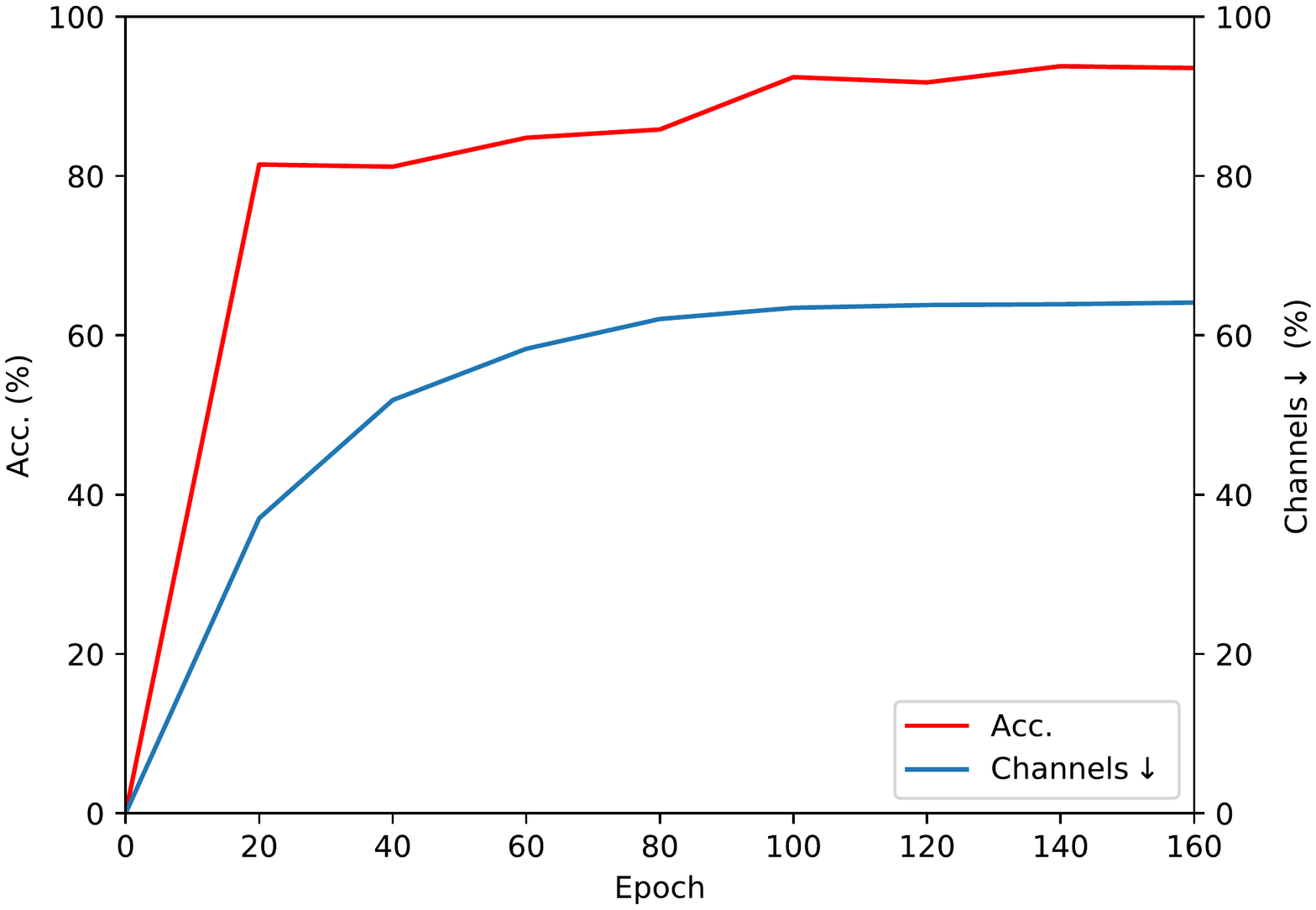}}
    \subfigure[ResNet-164]{\label{fig:4f}\includegraphics[width=42mm, trim=0 200 0 200, clip] {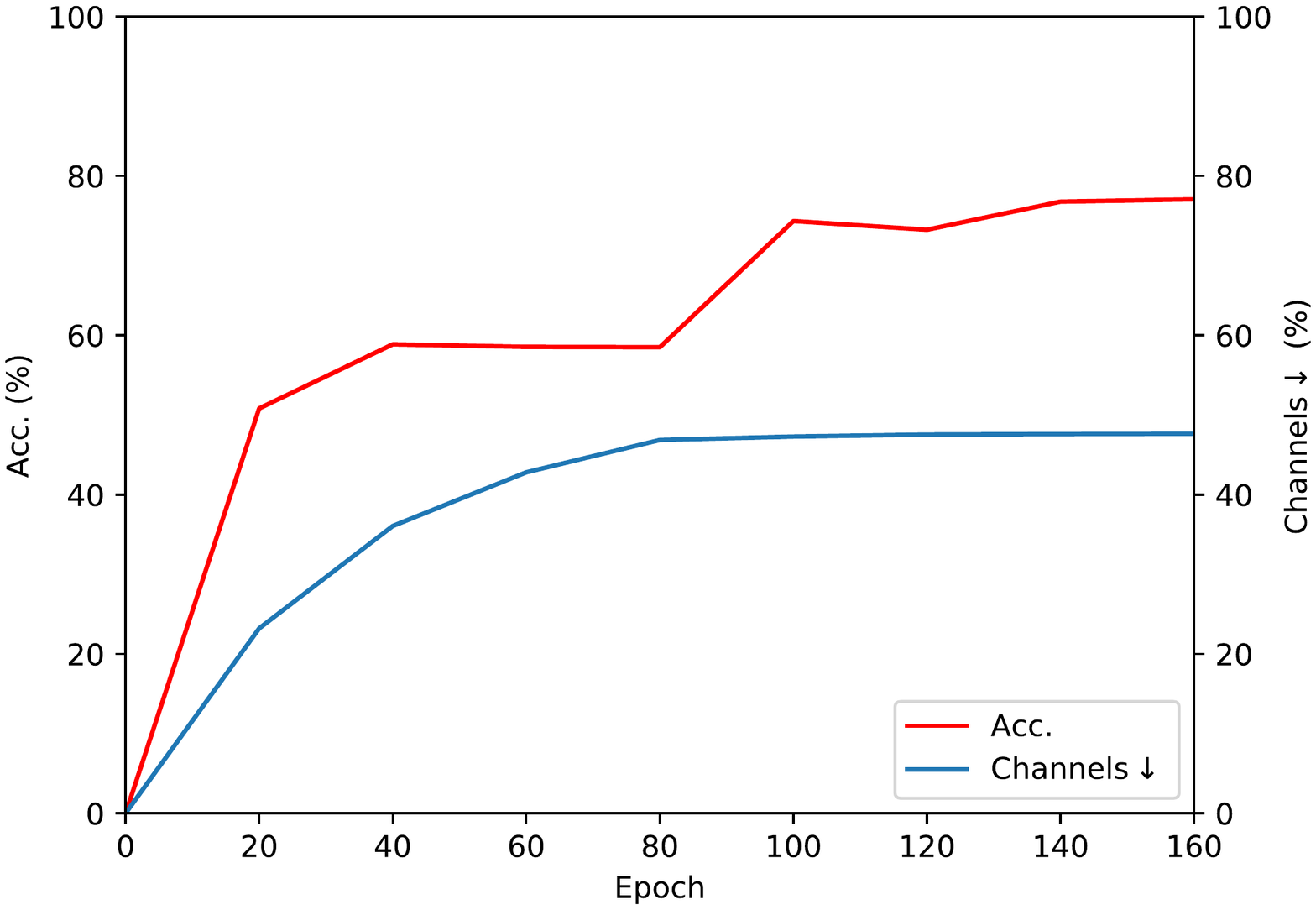}}
    \subfigure[DenseNet-40]{\label{fig:4e}\includegraphics[width=42mm, trim=0 200 0 200, clip] {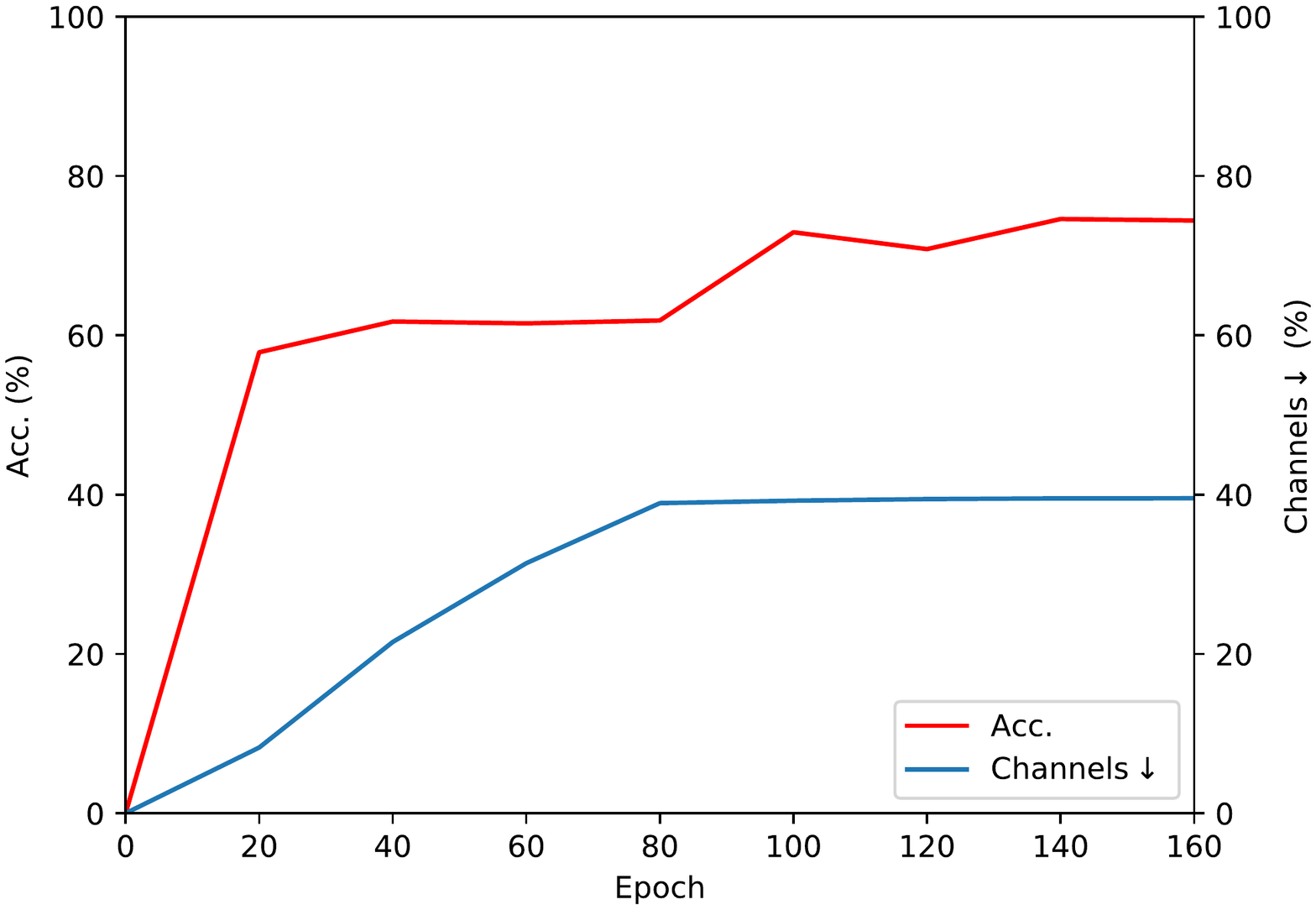}}
    \subfigure[VGGNet-19]{\label{fig:4g}\includegraphics[width=42mm, trim=0 200 0 200, clip] {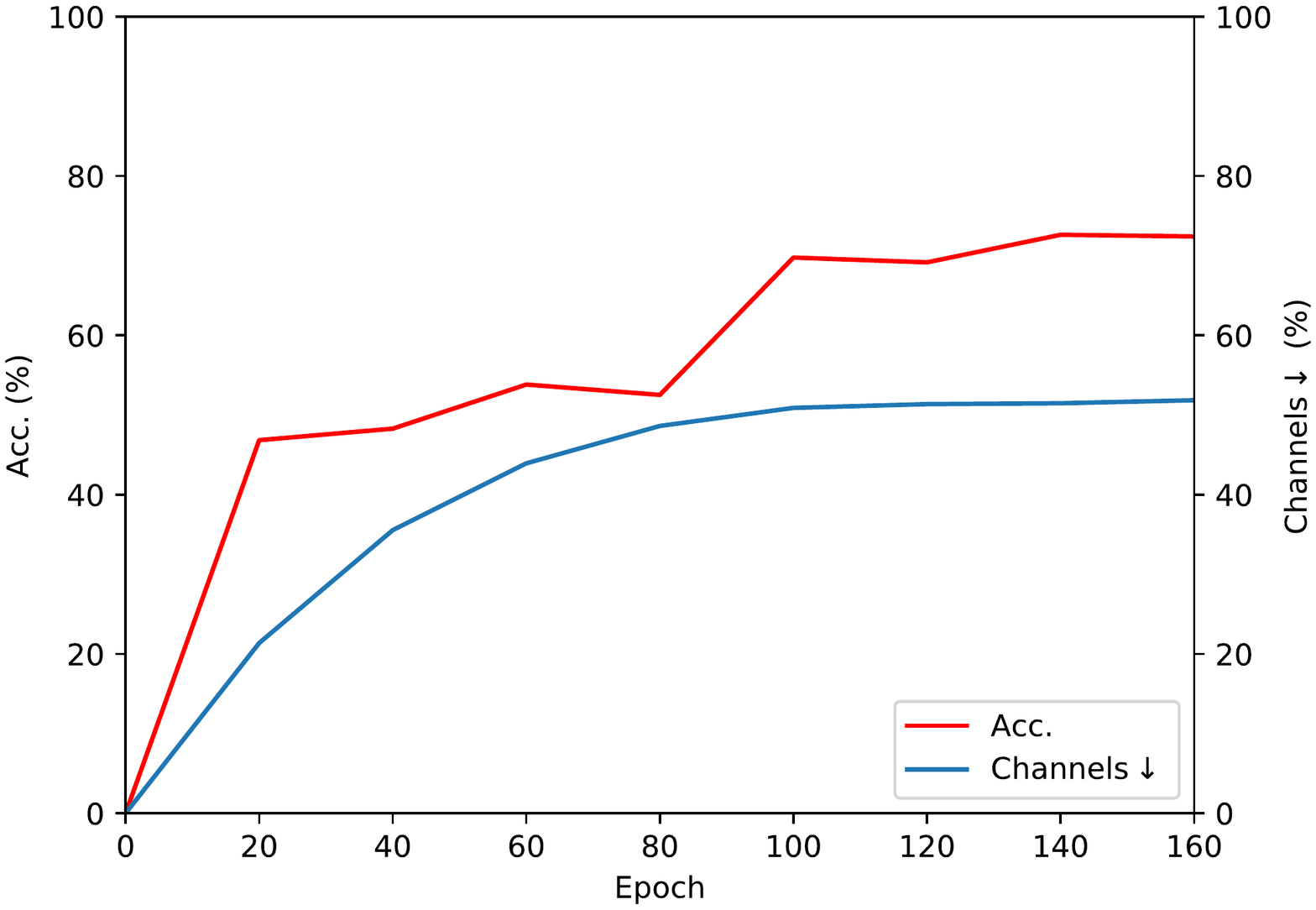}}
    \subfigure[VGGNet-16]{\label{fig:4h}\includegraphics[width=42mm, trim=0 200 0 200, clip] {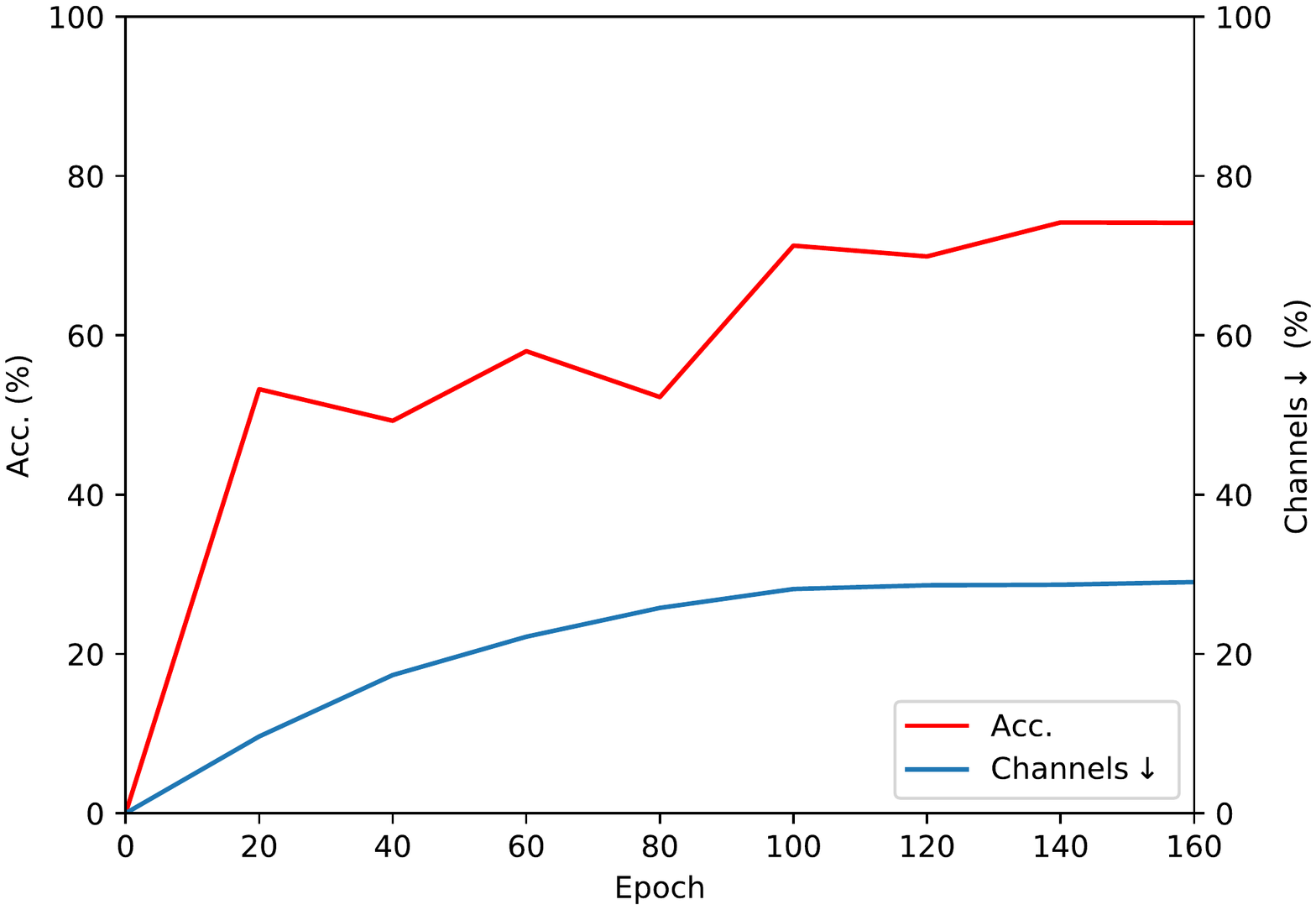}}
   \caption{Plots for accuracy and pruning ratio of channels versus epochs on CIFAR-10 (first row) and CIFAR-100 (second row) datasets using ResNet-164, DenseNet-40, VGGNet-19 and VGGNet-16 networks.
}
    \label{fig:plot_acc_channel}
\end{figure*} 
%

\paragraph{Effect of $\lambda$ for sparsity loss}
We analyze the effect of balancing term $\lambda$ for sparsity loss on CIFAR-100 using DenseNet-40 to investigate the trade-off between accuracy and FLOPs reduction.
Table~\ref{table:sensitivity_for_sparse} illustrates that the network becomes more compact as $\lambda$ increases.
Note that the network trained with $\lambda=1 \times 10^{-6}$ even outperforms the baseline model by 0.67\% points, which implies that our sparsity loss plays a role as a regularizer. 
\begin{table}[!t]
	\centering
	\caption{
		Sensitivity analysis about balancing term $\lambda$ for the sparsity loss ($\mathcal{L}_{\text{sparse}}$) using DenseNet-40 on CIFAR-100 dataset. 
	}
	\vspace{0.1in}
	\scalebox{0.80}{
		\begin{tabular}{c|c|c|c}
			\toprule
			 $\lambda$ &  Acc. (\%) & Acc. Drop & FLOPs $\downarrow$ (\%) \\
			\hline\hline	
			  $1 \times 10^{-6}$ & 74.91  & -0.67 & 28.99 \\
			  $5 \times 10^{-6}$ & 74.08  & 0.16  & 39.08 \\
			  $10 \times 10^{-6}$ & 73.84 & 0.40 & 46.25 \\
			   $30 \times 10^{-6}$ & 73.38  & 0.86  & 60.81 \\
			   $50 \times 10^{-6}$ & 73.17 & 1.07  & 67.82 \\
			\bottomrule			
		\end{tabular}
	}
	\label{table:sensitivity_for_sparse}
\end{table}
%

\paragraph{Effect of $s$ in confidence interval}
We study the effect of $s$ in \eqref{eq:sparsity_loss} on CIFAR-100 with DenseNet-40 to discuss accuracy and FLOPs trade-off.   
To this end, we test four different values of $s$, $\left\{ 0, 1, 2, 3 \right\}$, because the range induced by the values covers up to very high CDF values (more than 99.8\%) in the standard Gaussian distribution. 
Table~\ref{table:sensitivity_for_sparse_s} presents that larger values of $s$ lead to lower FLOPs but lower accuracies. 

\begin{table}[!t]
	\centering
	\caption{
		Sensitivity analysis about $s$ for the Gaussian confidence interval using DenseNet-40 on CIFAR-100 dataset with $\lambda=30 \times 10^{-6}$.
	}
	\vspace{0.1in}
	\scalebox{0.80}{
		\begin{tabular}{c|c|c|c}
			\toprule
			 $s$ &  Acc. (\%) & Acc. Drop & FLOPs $\downarrow$ (\%) \\
			\hline\hline	
			  0 & 74.59  & -0.35  & 45.73 \\
			  1 & 74.47  & -0.23 & 52.43 \\
			  2 & 73.86  & 0.38  & 58.45 \\
			  3 & 73.38 & 0.86 & 60.81 \\
			\bottomrule			
		\end{tabular}
	}
	\label{table:sensitivity_for_sparse_s}
\end{table}
\begin{table}[!t]
	\centering
	\caption{
		Performance comparison between our original algorithm and its modified version given the target pruning ratio 0.4 on ILSVRC-12 using ResNet-50. 
	}
	\vspace{0.1in}
	\scalebox{0.75}{
		\begin{tabular}{c|c|c|c|c}
			\toprule
			\multirow{2}{*}{Method} & \multirow{2}{*}{FT} & Top-1 & Top-5 & \multirow{2}{*}{FLOPs $\downarrow$ (\%)} \\
			& & Acc. Drop & Acc.Drop & \\
			\hline\hline	
			SCP & O & \textbf{0.49} & \textbf{0.53} & \textbf{49.9}\\
			SCP (modified training) & O & 0.58 & 0.57 & 48.9 \\
			\hdashline
			SCP & X & \textbf{1.33} & \textbf{0.92} & \textbf{49.9} \\
			SCP (modified training) & X & 1.43 & 1.04 & 48.9 \\
			\bottomrule
			
		\end{tabular}
	}
	\label{table:target_pruning}
\end{table}
%

\paragraph{Target channel pruning ratio}
The proposed algorithm needs to search for a proper value of $\lambda$ to control target pruning ratio, which is a drawback for its applicability.
However, such a limitation is addressed by a simple change of our training procedure.
We apply the sparsity loss to the channels with high CDF values (the ones within the target pruning ratio) and update the model based only on the sparsity loss when it increases; otherwise, the model is updated with the total joint loss for classification and sparsity.
To validate the effectiveness of this training scheme, we compare the model obtained from the new training strategy and the pruned network given by the original method with exhaustive search for $\lambda$ on ILSVRC-12 using ResNet-50.
Table~\ref{table:target_pruning} shows the new training algorithm achieves almost equivalent performance to the original version, even without time-consuming search for $\lambda$.

\paragraph{Balance between classification and sparsity loss}
Figure~\ref{fig:plot_acc_channel} illustrates both classification accuracy and pruning ratio with various backbone networks during their training procedures on CIFAR-10 and CIFAR-100 datasets.
Note that all cases show similar tendency;
1) both classification accuracy and network sparsity improve rapidly in the early stage of training,
2) the sparsity of networks is saturated after about 80 epochs,
and 3) the networks continue to enhance their accuracy even in the last part of training.

\section{Conclusion}
\label{sec:conclusion}
This paper presents a soft channel pruning algorithm by jointly learning model parameters and pruning masks via a stochastic gradient-descent method.
We formulate the channel pruning strategy in a principled way, based on the the property of a feature map derived by a sequence of BN and ReLU operations.
The proposed algorithm achieves outstanding performance in terms of accuracy and efficiency consistently even without extra fine-tuning, on the multiple standard benchmarks and over several different backbone networks.

\section*{Acknowledgements}

This work was partly supported by Samsung Advanced Institute of Technology (SAIT) and Institute for Information \& Communications Technology Promotion (IITP) grant funded by the Korea government (MSIT) [2016-0-00563, 2017-0-01779].

\nocite{langley00}

\bibliography{example_paper}
\bibliographystyle{icml2020}

\end{document}